\pdfoutput=1

\documentclass[11pt]{article}

\usepackage{acl}

\usepackage{times}
\usepackage{latexsym}
\usepackage{adjustbox}
\usepackage{paralist}
\usepackage{amsmath}
\usepackage{enumitem}
\usepackage{array}
\usepackage[T1]{fontenc}

\usepackage[utf8]{inputenc}
\makeatletter
\newcommand{\printfnsymbol}[1]{%
  \textsuperscript{\@fnsymbol{#1}}%
}
\makeatother
\usepackage{microtype}

\usepackage{fancyhdr}
\title{CDialog: A Multi-turn Covid-19 Conversation Dataset for Entity-Aware Dialog Generation}

 \author{Deeksha Varshney$^\dagger$\thanks{\hspace{4pt}Equal Conrtibution}\hspace{4pt}, \quad Aizan Zafar$^\dagger$\printfnsymbol{1}, \quad Niranshu Kumar Behra$^\dagger$ \quad Asif Ekbal$^\dagger$\\
$^\dagger$Department of Computer Science and Engineering,\\ Indian Institute of Technology Patna, India\\
{\tt \small $\{$1821cs13,aizan\_1921cs17,niranshu\_1901cs39,asif$\}$@iitp.ac.in} \\
}

\begin{document}
\maketitle
\begin{abstract}
The development of conversational agents to interact with patients and deliver clinical advice has attracted the interest of many researchers, particularly in light of the COVID-19 pandemic. The training of an end-to-end neural based dialog system, on the other hand, is hampered by a lack of multi-turn medical dialog corpus. We make the very first attempt to release a high-quality multi-turn Medical Dialog dataset relating to Covid-19 disease named \textit{CDialog}, with over 1K conversations collected from the online medical counselling websites. We annotate each utterance of the conversation with seven different categories of medical entities, including diseases, symptoms, medical tests, medical history, remedies, medications and other aspects as additional labels. Finally, we propose a novel neural medical dialog system based on the CDialog dataset to advance future research on developing automated medical dialog systems. We use pre-trained language models for dialogue generation, incorporating annotated medical entities, to generate a virtual doctor's response that addresses the patient's query. Experimental results show that the proposed dialog models perform comparably better when supplemented with entity information and hence can improve the response quality. 

\end{abstract}

\section{Introduction}
Currently, telemedicine is absolutely appropriate in reducing the risk of COVID-19 among healthcare providers and patients due to the diversion of medical resources as millions of people around the world have experienced delays in diagnosis and treatment. Conversational agents \cite{gopalakrishnan2019topical,zhao2020pre,wu2018global,reddy2019multi} have been proved to be effective in carrying on a natural conversation and understanding the meanings of words to respond with a coherent dialog. It has been also effective in providing support to complete several tasks such as booking a ticket \cite{liao2020task}, getting reservations \cite{wei2018task}, etc. In medical domain, \cite{zeng2020meddialog,liu2020meddg,li2021semi,wei2018task,xu2019end} have come up with standard techniques to model medical dialogs which reduces face-to-face consultations, resulting in reduced costs and helps the patient get quicker medical treatment. However, medical dialog systems are more difficult to implement than the standard task-oriented dialog systems (TDSs) as there are several other professional phrases / formal medical expressions that are frequently conveyed while communicating \cite{shi2020understanding}. 

\begin{figure*}[ht!]
    \centering
    \includegraphics[width=0.98\textwidth]{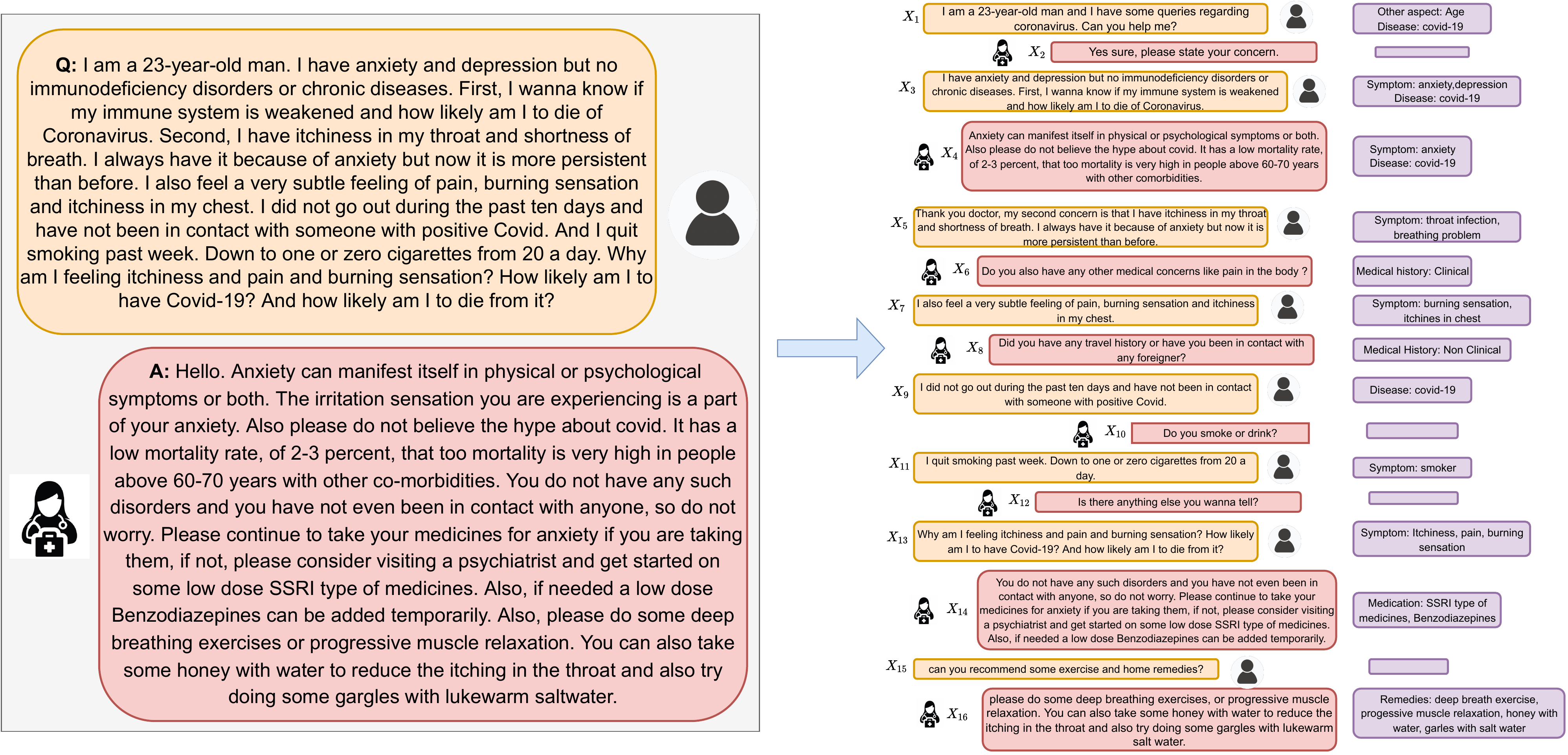}
    \caption{{Sample conversation from the CDialog dataset. Sample on left side is from existing CovidDialog dataset \cite{yang2020generation}. We have extended this to a multi-turn dialog with eight turns along with entity information. Right side shows such extended samples.}}
    \label{fig:example}
\end{figure*}

A significant effort has recently been undertaken to collect medical dialog data for research on medical dialog systems \cite{shi2020understanding,zeng2020meddialog,liao2020task,liu2020meddg,yang2020generation}. They all, however, have some limitations:
\begin{inparaenum}[(i)] 
\item A comprehensive diagnosis and treatment procedure is lacking.
\item Labels are not fine-grained enough. Prior research has typically provided a single poorly graded label for the entire utterance, which may mislead model training and/or lead to erroneous assessment. Furthermore, the scale of the medical entities involved is limited.
\item Dialog length is limited to an average of 2 turns only.
\end{inparaenum}
{From Figure \ref{fig:example}, it can be seen that the original CovidDialog corpus \cite{yang2020generation} has a dialog with only one turn and the patient and doctors utterances are also too lengthy having all the information together at one place. We attempt to split this dialog to make it more suitable for dialog settings by separating and pairing the doctors' and patients' utterances at appropriate points. 
For example, the first sentence of the patient's query (c.f $Q$) from Figure \ref{fig:example}, is chosen as the first utterance (c.f $X_1$) for the multi-turn dialog as shown on the right. To maintain the dialog flow, we include generic utterances  by doctors as the second utterance such as \textit{``Yes sure, please state your concern."} (c.f $X_2$). We also include appropriate sentences from doctor's response (c.f $A$), as subsequent utterances (c.f $X_4$) which comprehends to patient's utterance (c.f $X_3$) at that point.}

{Further, we also assign fine-grained medically relevant categories to these utterances. For example, for the third utterance in Figure \ref{fig:example}, there are two different kinds of categories: informing symptom status \textit{(Symptoms: anxiety, depression)} and inquiring diseases \textit{(Disease: Covid-19)}.}

To address the issue of lack of medically relevant dialog data, we create \textit{CDialog}, a multi-turn Medical Dialog dataset pertaining to Covid-19 disease. As indicated in Table \ref{tab:comp_dial}, our dataset has the following advantages over the existing conversational datasets. First, our dataset is the largest Covid-19 related dialogue dataset with highest average number of dialogue turns, 
and thus more suitable for training neural conversation models. Second, CDialog is informational and diversified, with 12 types of diseases and 253 types of entities, which is far more representative of an actual medical consultation scenario. Furthermore, to gain a better grasp of the response generation task, we compare a number of cutting-edge models on CDialog by using popular pre-trained language models like BERT \cite{devlin-etal-2019-bert} and GPT \cite{radford2019language}. Moreover, we create a medical entity-aware dialog system that makes use of entity-level knowledge. According to the experimental results, combining entity information with dialog history in the generation process improves the response quality.

Our current work makes the following contributions:
\begin{enumerate}[nolistsep]
\item We build and release \textit{CDialog}, a multi-turn medical dialog dataset related to Covid-19. CDialog has around 1K conversations and with more than 7K utterances annotated with seven types of medical entities, giving it a credible standard for evaluating the medical consultation capabilities of dialog systems.
\item On the CDialog dataset, we present several baselines for response generation and propose techniques for utilizing the relevant medical dialog entities in the medical dialog system.
\item We conduct rigorous experiments, including quantitative and qualitative evaluation, to evaluate a number of cutting-edge pre-trained models for medical dialog generation. Empirical evaluation demonstrates 
that annotated entities as auxiliary information significantly improves the response quality.
\end{enumerate}

\begin{table*}[t!]
\begin{center}
\small
\begin{adjustbox}{max width=1.0\textwidth}
\renewcommand{\arraystretch}{1.2}
\setlength\tabcolsep{1.2pt}
\begin{tabular}{|>{\centering\arraybackslash}m{5.0cm}|>{\centering\arraybackslash}m{3.9cm}>{\centering\arraybackslash}m{1.8cm}>{\centering\arraybackslash}m{1.8cm}>{\centering\arraybackslash}m{1.8cm}>{\centering\arraybackslash}m{2.9cm}>{\centering\arraybackslash}m{1.8cm}|}
\hline
\textbf{Dataset} & \textbf{\#Domain} & \textbf{\#Diseases} & \textbf{\#Dialogs} & \textbf{\#Utterances} & \textbf{Avg. Dialog length} & \textbf{\#Entities}  \\ 

\hline

DX(Dxy) \cite{xu2019end} & Pediatrics  & 5  & 527  & 2,816 & 5.26 &  46   \\

COVID-EN \cite{yang2020generation}  & COVID-19   & 1   & 603         & 1,232      & 2.00 & -    \\

MedDialog-EN \cite{zeng2020meddialog} & Diabetes, elderly problems, pain management, etc \footnote{Only }   & 96 & 260,000   & 510,000    & 2.00 & -    \\

\hline

\textbf{CDialog (ours)} & \textbf{COVID-19 \& related symptoms}    & \textbf{12}  & \textbf{1,012}   & \textbf{7,982}  & \textbf{8.00} & \textbf{253}    \\
\hline
\end{tabular}
\end{adjustbox}
\end{center}
\caption{Comparison of our corpus to other medical dialog corpora. Statistics include the number of dialogs, disease types, utterances, entity types and average dialog length.}
\label{tab:comp_dial}
\end{table*}

\section{Related Work}
\label{sec:related_works}
For dialog generation, sequence-to-sequence models  \cite{vinyals2015neural,sutskever2014sequence} are very popular. \citet{shang2015neural} proposed a recurrent neural network (RNN) based encoder-decoder architecture for short text conversations. \citet{li2016diversity,xing2017topic,zhao2017learning,tao2018get} developed models to help improve the performance of traditional dialog systems using extra features such as topic of the conversation, different objective function. 
\citet{serban2016building,serban2017hierarchical,xing2017topic,zhang2019recosa} proposed a number of models for efficiently selecting the conversational context in multi-turn conversation system.

Recent work by \cite{zhang2020dialogpt} using pre-trained language models has demonstrated captivating performance on generating responses that make sense under the conversation contexts while also carrying out specific content to keep the conversation going by fine-tuning GPT-2 \cite{radford2019language} in different sizes on social media data. Among all accessible pre-trained language models, BERT is commonly utilised in the medical domain, as several models, such as BioBERT \cite{lee2020biobert}, Clinical-BERT \cite{alsentzer2019publicly}, and so on are implemented using the data from a specific domain.

Information extraction \cite{zhang2020mie}, relation prediction \cite{du2019learning, lin2019enhancing,xia2021medical}, and slot filling \cite{shi2020understanding} are some of the recent tasks performed on medical data. In medical domain, the use of a reinforcement learning framework in dialog systems \cite{wei2018task} has encouraged dialog management strategy learning. Further \cite{xu2019end} increased the rationality of medical conversation decision-making by including external probabilistic symptoms into a reinforcement learning framework. \citet{liao2020task,xia2020generative} used hierarchical reinforcement learning for automatic disease diagnosis. These RL systems, on the other hand, solely learn from tabular data containing the existence of symptoms, ignoring the importance of other key information such as symptom features, tests, and treatment. Furthermore, \cite{ferguson2009cardiac,wong2011health,gatius2012conversational,liu2016augmented} constructed early end-to-end medical dialog systems on large scale Chinese medical dialog corpora. 


\begin{figure}[ht!]
    \centering
    \includegraphics[width=0.50\textwidth]{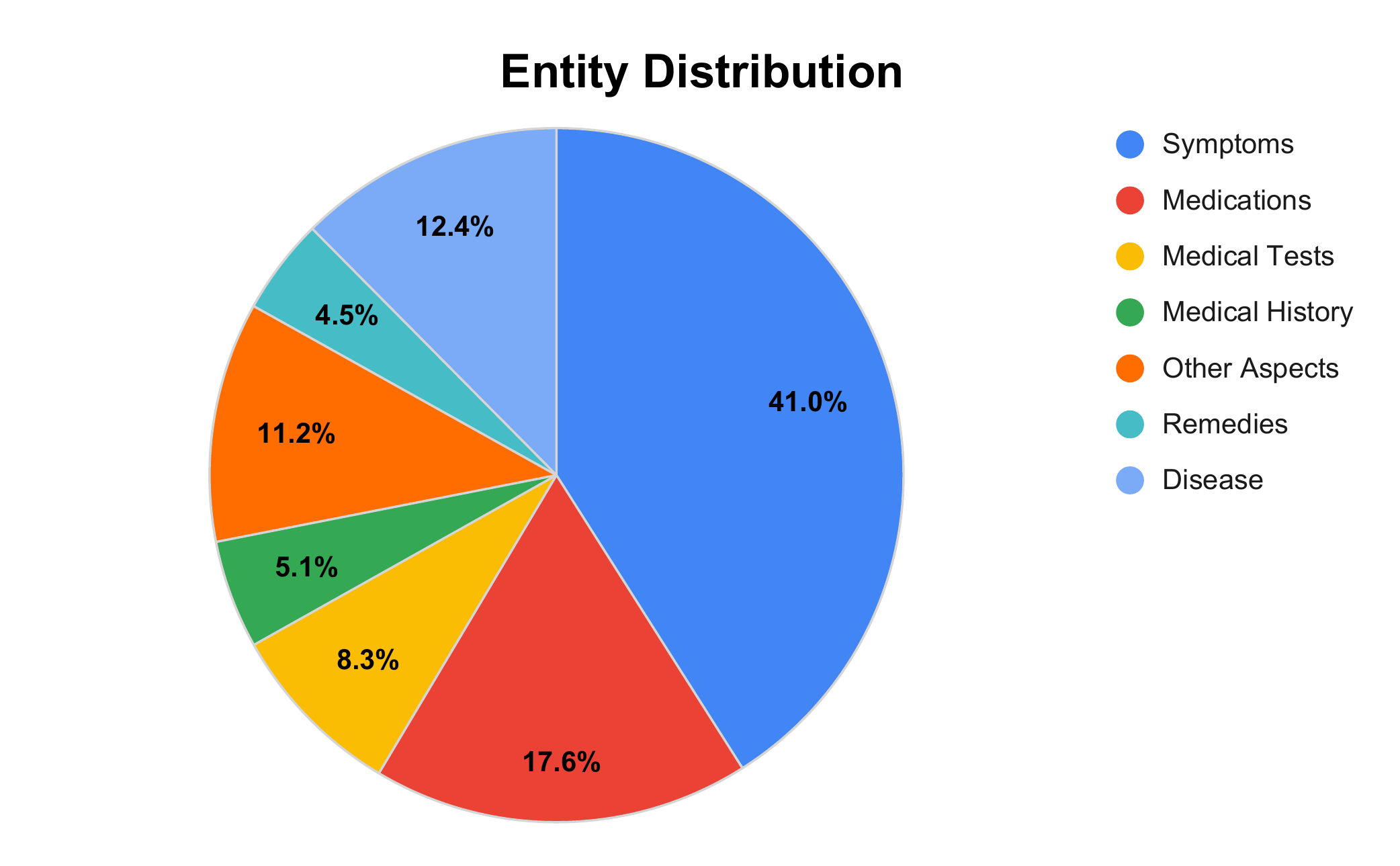}
    \caption{Entity distribution in the CDialog dataset}
    \label{fig:entity_dist}
\end{figure}

\citet{wei2018task} released the first dataset for medical diagnosis, although it only includes structured user goal data rather than natural language dialog. \citet{xu2019end} released a simple dataset named DX with 527 real language dialogs.
Recently, \cite{zeng2020meddialog} released a high-quality unlabelled medical dialogue dataset named MedDialog in Chinese and English covering more than 50 diseases. Although, MedDialog corpora contains the highest number of dialogs, they do not cover dialogs on Covid-19 and have an average dialogue length of only 2. Furthermore, \cite{shi2020understanding} released a general-domain medical dialog corpus containing 2K labelled data and 100K unlabeled data, but in the form of individual utterances rather than the entire dialog. MedDG \cite{liu2020meddg} compared to the previous corpora involved more diseases, entities, dialogs, and the utterances to alleviate the issue of data scarcity. \citet{li2021semi} also released a high quality  knowledge-aware medical conversation dataset (KaMed)
from ChunyuDoctor, a large online Chinese medical consultation platform. Similar to previous datasets, \cite{shi2020understanding,liu2020meddg,li2021semi} did not focus on Covid-19 disease.

We create and release a multi-turn dialog dataset named \textit{CDialog} which contains 1K English consultations between patients and doctors along with medical entity annotated utterances. Finally, we propose an entity-aware neural medical conversation model that generates appropriate responses by utilizing the annotated entities.

\section{Resource Creation}
In this section, we describe the details of resource creation. 
\subsection{CDialog Dataset}
We extend the CovidDialog dataset \cite{yang2020generation} with the dialogs from the diseases which are the symptoms of Covid-19 and named it as Ext-CovidDialog which now contains approximately 10K dialogs. The motivation for extending the dataset comes from the fact that a conversation about Covid-19 can benefit from the conversations about fever, cough, cold, and other symptoms of Covid-19. We used online platforms of health service consultations such as icliniq.com and heathcaremagic.com to crawl data for fever, cough, etc. We extended the dialog length of 1K dialogs (from 2 to 8) using the dialogs from Ext-CovidDialog (contains $\sim$ 10K dialogs) and also annotated them with several medical entities. The resulting dataset is named as \textit{CDialog} which is finally our proposed dataset for this work.

Our motivation is within the scope of building a conversational system that would engage in online conversation with the users. While developing an automated conversational system, generating longer responses is often a problem for the deep learning models. Hence, we have manually broken this longer utterance into multiple turns. We interacted with the medical experts in our university hospital to ensure that such splitting does not distort the crucial health-related information, rather we added generic statements in order to maintain the flow of the conversation.

\subsubsection{Construction Details}
\label{sec:const_det}
As shown in Figure  \ref{fig:example}, we show a sample of the created and annotated conversation from the CDialog dataset. The average number of utterances in the crawled data (Ext-CovidDialog) is 2.0 per conversation, and the average number of tokens in an utterance is 103. As a result, this conversation is more akin to a question-and-answer session, with the patient describing their problem in detail and the doctor thoroughly answering each question. We aim to convert this question-and-answer (c.f Figure \ref{fig:example} left) setup into a multi-turn human-like conversation format (c.f Figure \ref{fig:example} right). For this, we first view the patient query (c.f $Q$ in Figure \ref{fig:example}) as a combination of individual sentences such that each sentence represents some meaningful intent. Then, we choose an appropriate sentence to start the conversation. For each chosen sentence from the patient's query, we search for its significant response in the doctor's answer (c.f $A$ in Figure \ref{fig:example}). We have introduced/modified the dialogs in between as needed to ensure that all dialogs are continuously readable and do not go out of context. {Because medical data annotation involves annotators with proficient medical knowledge, the annotation cost is high. We employ four annotators with relevant medical expertise. Before beginning the annotation process, we explained the annotation guidelines (c.f Appendix \ref{sec:annotation_guide}) using a few examples from the dataset to the annotators. 
We observe a Fleiss’ kappa \cite{fleiss1971measuring} score of 0.85 among the annotators denoting good agreement between them for the task of converting single turn dialogs into multi turn dialogs.
}

\paragraph{Medical Entity Annotation:} We choose the following seven different kinds of entities for annotation after consulting with domain experts: \textit{Diseases} such as allergic conjunctivitis, allergic cough, bacterial conjunctivitis, and so forth; \textit{Symptoms} such as pneumonia, body ache, cough and so on; \textit{Medication} such as anti-allergic tablets, betadine gargle solution, hydroxychloroquine and so on; \textit{Medical Tests}, such as x-rays, etc; \textit{Medical history}, which may be ``clinical" or ``non-clinical"; \textit{Remedies} such as gargle, exercise, and so on; and \textit{other factors} such as age, nature of pain, duration, and location. 
As a result, we have 253 entities consisting of 25 different medical tests, 87 different symptoms, 138 different medications, 12 different diseases, 2 different medical histories, 10 unique remedies and 4 other aspects. The distribution of entities in the CDialog dataset is depicted in Figure \ref{fig:entity_dist}. It shows the proportion of entities in each of the seven categories. Each utterance of the conversation is labeled separately using the seven entity categories, as shown in the right side of Figure 1. {The annotation process involved four annotators with relevant medical backgrounds. They begin by discussing the creation of an annotation template. Each participant annotates a small portion of the data and reports the confusing utterance. We summarize our observations and then revise the annotations once more. We observe a Fleiss’ kappa \cite{fleiss1971measuring} score of 0.89 between annotators denoting great agreement between them for the entity annotation task.}

More details on the platform and annotators payment can be found in the Appendix \ref{sec:annotation_guide}.

\subsubsection{Dataset Statistics and Comparision to Existing Dataset}
As a result of the annotation process as described in Section \ref{sec:const_det}, the \textit{CDialog} dataset contains 1012 English consultations about Covid-19 and Covid-related symptoms, such as allergic conjunctivitis, allergic cough, bacterial conjunctivitis, and so forth, which aids in building the multi-turn dialog generation model. The total amount of tokens is 1,085,204 and the total number of utterances is 7,982. The average, maximum, and minimum number of utterances are 8.0, 48, and 2, respectively. The average, maximum, and minimum number of tokens in an utterance are 136, 5313, and 2, respectively. The dataset statistics is shown in Table \ref{tab:dataset_stats} in the Appendix \ref{sec:data_details}.

We compare our proposed \textit{CDialog} dataset to the other publicly available datasets in Table \ref{tab:comp_dial} and observe that only three out of the many available datasets as mentioned in Section \ref{sec:related_works} are in English. When compared to these datasets, we find that the average dialogue length in \textit{CDialog} is eight, indicating that it is more conversational in nature, and our dataset is the largest, focusing solely on Covid-19 with entity annotation for developing entity-aware language models.

\section{Methodology}
\subsection{Task Definition}
\begin{figure}[ht!]
    \centering
    \includegraphics[width=.50\textwidth]{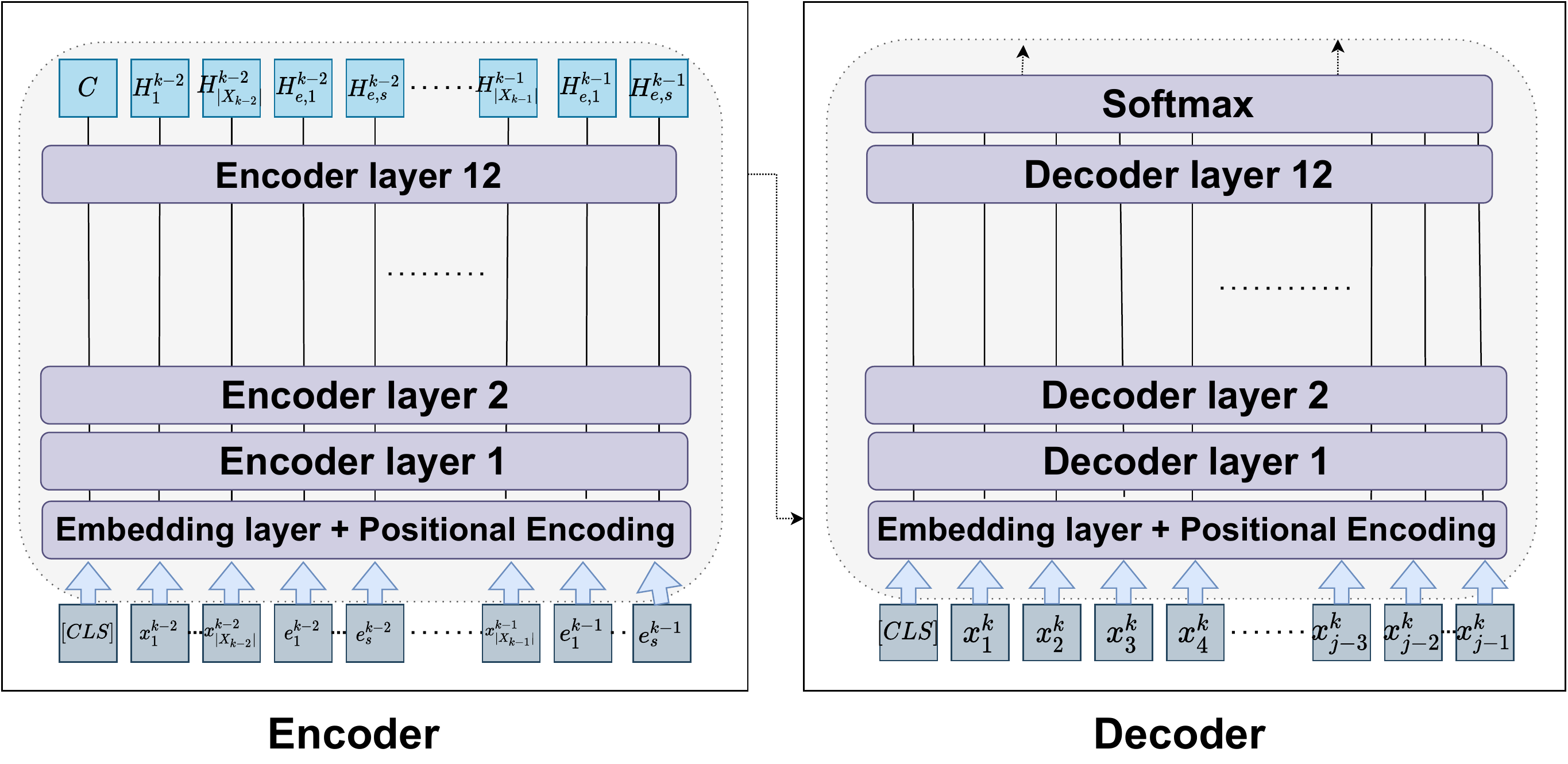}
    \caption{Model architecture}
    \label{fig:model}
\end{figure}

The goal of a medical dialog system is to provide context-consistent and medically inclined responses based on conversation histories. Formally, given the history of conversations between doctor and patient comprising of K utterances, $X = {X_1, X_2, ..., X_i, .., X_K}$,  where $X_i$ is either a doctor's or a patient's utterance. Each utterance is tagged with an entity set $E = {e^1_1 , ..., e^1_s, ... e^K_1 , ..., e^K_s }$, where $s$ is the total number of entities associated with an utterance, $X_i$. The response generation task is to generate $Y = {y_1, y_2, ..., y_M }$ with $M$ words given the set of previous $K$ utterances with entity set $e^K_s$. The architecture is shown in Figure \ref{fig:model}. 

\subsection{Entity-aware Dialog Model}
Since generative models are inapplicable to our dataset's annotated entity labels, we present entity-aware models that make use of the supplementary entity knowledge. In this method, the entity set after the dialog history is directly concatenated as new input text and then used to encourage the models for generating the relevant responses.

\subsubsection{Model Description}
To generate contextualized utterance representation for the input sequences, we use the BioBERT\_BASE \cite{lee2020biobert} pre-trained model (Cased: hidden-1024, heads-16, layer-8, 1M parameters). The context utterances are concatenated with the current user utterance to form a single input utterance. The following is the flattened token sequence for the input utterance combined with the associated entity set:

\begin{multline}
[CLS], x^{k-2}_1, ..., x^{k-2}_{|X_{k-2}|}, e^{k-2}_1, ..., e^{k-2}_s, \\ [SEP], x^{k-1}_1, ..., x^{k-1}_{|X_{k-1}|}, e^{k-1}_1, ..., e^{k-1}_s [SEP]
\end{multline}

where the $[CLS]$ token is inserted at the start of the sequence to indicate the beginning of the sentence. The $[SEP]$ token denotes the end of a sentence and distinguishes one sequence from the next. Each token is first embedded through three layers (Token, Segment, and Position). The hidden states are obtained by feeding the respective vectors obtained from these three embedding layers into the BioBERT encoder. Furthermore, the hidden vector for each $i$-th word in the input utterance is denoted as $H^{k-1}_{i}$. The bidirectional nature of BioBERT ensures joint conditioning on both the left and right contexts of a token. Then, using a BioBERT decoder, we generate the doctor's response, $Y = {y_1, y_2, ..., y_M }$, using the words from the gold response, $X_k$ = ($x^k_{1}$,$x^k_{2}$,.....,$x^{k}_{|X_k|}$) every time. The decoder predicts each word, $y_j$, conditioned on $x^k_{1}$,...,$x^k_{j-1}$, $H^{1}_{1}, ..., H^{k-1}_{1},...,H^{k-1}_{{|H_{k-1}|}},H^{k-1}_{{e,1}}, ..., H^{k-1}_{{e,s}}$
\begin{gather}
\label{eq:9}
    Q_j^k = BioBERT\_decoder(H^{k}) \\
    P(y^k_j) = \text{softmax}(Q_j^k)
\end{gather}

\subsubsection{Training Loss}
The decoder loss is the cross-entropy between the output distribution $P(y^k_j)$ and the reference distribution, $T_j$, denoted as
\begin{gather}
    Loss = - {\sum}{T_j}log(P(y^k_j)))
\end{gather}

\section{Experimental Setup}
This section describes the baseline models and evaluation metrics. Implementation details can be found in the Appendix \ref{sec:imp_det}.

\subsection{Baselines}
\label{sec:sota}
We use the following baseline models: 

\begin{inparaenum}

\item \textbf{$\text{GPT-2}$ \cite{radford2019language}}: It is a language model based on Transformer pretrained on Reddit dialogs, in which the input sequence is passed through the model to generate conditional probability on the output sequences.

\item \textbf{$\text{DialogGPT}_{\text{\textit{finetune}}}$ \cite{zhang2020dialogpt}}: 
The model was trained using 147 million Reddit chats and is based on the OpenAI GPT-2 architecture. We begin by concatenating all dialog turns within a dialogue session into a long text that is terminated by the end-of-text token.

\item \textbf{BERT \cite{devlin2018bert}}: 
This model makes use of Transformer attention mechanism which learns contextual relations between the words (or, sub-words) in a text. BERT as an encoder is used to encode the input and BERT as a decoder is used to generate relevant output. 

\item \textbf{BART \cite{lewis2019bart}:} 
In this model a bidirectional encoder is used for encoding the input sequences and the appropriate response is generated using a left-to-right decoder.

\item \textbf{BioBERT \cite{lee2020biobert}}: 
BioBERT is a model similar to BERT aside from that it has been pre-trained on a large biomedical corpus. It outperformed BERT and other state-of-the-art models in several tasks of biomedical text analysis. We use BioBERT both as the encoder and decoder. 

{The entity set after the dialogue history is directly concatenated as new input text in \textbf{BERT-Entity, BART-Entity, and BioBERT-Entity} and then used to stimulate the models to produce the relevant responses.}

\end{inparaenum}

\begin{table*}[t!]
\begin{center}
\small
\begin{adjustbox}{max width=1.0\textwidth}
\renewcommand{\arraystretch}{1.3}
\setlength\tabcolsep{1.2pt}
\begin{tabular}{|>{\centering\arraybackslash}m{3.0cm}|>{\centering\arraybackslash}m{1.9cm}|>{\centering\arraybackslash}m{1.9cm}|>{\centering\arraybackslash}m{2.0cm}|>{\centering\arraybackslash}m{2.0cm}|>{\centering\arraybackslash}m{2.0cm}|>{\centering\arraybackslash}m{2.0cm}|>{\centering\arraybackslash}m{2.0cm}|>{\centering\arraybackslash}m{2.0cm}|>{\centering\arraybackslash}m{2.0cm}|>{\centering\arraybackslash}m{1.6cm}|>{\centering\arraybackslash}m{1.6cm}|>{\centering\arraybackslash}m{1.8cm}|>{\centering\arraybackslash}m{1.8cm}|>{\centering\arraybackslash}m{1.8cm}|>{\centering\arraybackslash}m{1.8cm}|>{\centering\arraybackslash}m{2.0cm}|>{\centering\arraybackslash}m{1.8cm}|>{\centering\arraybackslash}m{1.8cm}|}
\hline
\textbf{Models} & \textbf{PPL }  & \textbf{F1\% } & 
 \textbf{BLEU-1 } & \textbf{BLEU-2 } & \textbf{BLEU-3 } & \textbf{BLEU-4 } & \textbf{ROUGE-L } & \textbf{Embedding Average} & \textbf{Vector Extrema
} & \textbf{Greedy Matching} \\
\hline

$\text{GPT-2}$ & {55.45 } & {9.43} & 
{0.145} &   {0.044} &  {0.018} &  {0.009} &  {0.108} &  {0.820} &  {0.355} &   {0.630}  \\

$\text{DialogGPT}_{\text{\textit{finetune}}}$ & {52.34 } & {9.89} & {0.148} &   {0.048} &  {0.019} &  {0.009} &  {0.109} &  {0.832} & 
{0.359} &   {0.637}  \\

BERT & {38.48} & {10.01} & 
{0.147} &   {0.045} &  {0.021} &  {0.012} &  {0.124} &  {0.851} & 
{0.360} &   {0.640} \\

BART & {25.14} & {11.82} & 
{0.161} &   {0.059} &  {0.029} &  {0.017} &  {0.139} &  {0.855} & 
{0.368} &   {0.644} \\

BioBERT & \textbf{22.67} & {15.68} & 
{0.204} &   {0.100} &  {0.066} &  {0.051} &  {0.174} &  {0.862} & 
{0.401} &   {0.663} \\

\hline
BERT-Entity & {38.40} & {10.36} & 
{0.150} &   {0.049} &  {0.020} &  {0.010} &  {0.123} &  {0.849} & 
{0.353} &   {0.637} \\

BART-Entity & {25.92} & {11.81} & 
{0.168} &   {0.061} &  {0.032} &  {0.020} &  {0.138} &  {0.854} & 
{0.362} &   {0.643} \\

BioBERT-Entity & {22.97} & \textbf{17.60} & 
\textbf{0.217} &   \textbf{0.126} &  \textbf{0.094} &  \textbf{0.078} &  \textbf{0.191} &  \textbf{0.865} & \textbf{0.404} &   \textbf{0.667} \\

\hline
\end{tabular}
\end{adjustbox}
\end{center}

\caption{\label{tab:auto_results}
{Automatic evaluation results for the baseline and suggested model on CDialog  dataset. BERT-Entity, BART-Entity, and BioBERT-Entity: BERT, BART and BioBERT based models with the entities concatenated with the input sequences, respectively.}}

\end{table*}

\begin{table}[ht!]
\begin{center}
\small
\begin{adjustbox}{max width=0.40\textwidth}
\renewcommand{\arraystretch}{1.1}
\setlength\tabcolsep{1.0pt}
\begin{tabular}{|>{\centering\arraybackslash}m{2.3cm}|>{\centering\arraybackslash}m{1.0cm}|>{\centering\arraybackslash}m{1.3cm}|>{\centering\arraybackslash}m{1.4cm}|>{\centering\arraybackslash}m{1.0cm}|>{\centering\arraybackslash}m{2.0cm}|>{\centering\arraybackslash}m{2.0cm}|>{\centering\arraybackslash}m{2.0cm}|>{\centering\arraybackslash}m{2.0cm}|}
\hline
\textbf{Models} & \textbf{Fluency}  & \textbf{Adequacy} &  \textbf{Entity Relevance} &  \textbf{Kappa} \\
\hline
BERT &  {2.65}   &  {1.80} & {2.59}   &  {0.87}    \\
BART &  {3.31}   &  {2.18} & {1.92}   &  {0.86}      \\
BioBERT &  {3.60}   &  {2.31} & {2.00}   &  {0.85}      \\   
\hline
BERT-Entity  &  {2.71}   &  {1.84} & {1.69}   &  {0.81}   \\
BART-Entity  &  {3.16}   &  {2.33}  & {2.06}   &  {0.82}    \\
BioBERT-Entity &  \textbf{3.55}   &  \textbf{2.86} & \textbf{2.33}   &  {0.82}      \\

\hline
\end{tabular}
\end{adjustbox}
\end{center}

\caption{\label{tab:h_results}
Human assessment results for the baseline and proposed model on the CDialog datasets. The bolded values represent the best value.}
\end{table}

\subsection{Evaluation Metrics}
\subsubsection{Automatic Evaluation} We evaluate our models on test set, using the following standard metrics. The BLEU \cite{papineni2002bleu} score computes the amount of word overlap with the words from the ground truth response. ROUGE-L \cite{lin-2004-rouge} measures the longest matching sequence of words between the candidate and the reference summary using longest common sub sequence method. Perplexity (PPL) is computed to learn how well the system learns to model the dialog data. We also compute \textit{unigram} F1-score \footnote{https://github.com/facebookresearch/ParlAI/blob/master/
parlai/core/metrics.py} between the predicted sentences and the ground truth sentences. Embedding-based metrics \footnote{https://github.com/Maluuba/nlg-eval} \cite{liu2016not} such as Greedy Matching, Vector Extrema and Embedding Average are an alternative to word-matching-based metrics. These metrics assign a vector to each word in order to comprehend the desired sense of the predicted sentence, as described by the word embedding.

\subsubsection{Human Evaluation}To evaluate the quality of generated responses from a human point of view, we randomly select 50 dialogs from each model developed using the CDialog dataset and analyze the predicted responses with the assistance of three human evaluators. For each example, we provide the responses (generated by models and ground-truth by humans) to our annotators. Human raters are post-graduates in science and linguistics with annotation experience for text mining tasks. We also had our model outputs validated by a doctor with a postgraduate degree in medicine. The important medical information was found to be retained in the responses. To assess the accuracy of our model predictions, we employ the following metrics: \begin{inparaenum}[(i)] 
\item Fluency: It is a measure of sentence's grammatical correctness.
\item Adequacy: This metric is used to determine whether the generated response is meaningful and relevant to the conversation history.
\item Entity Relevance (ER): This metric is used to determine whether or not a response contains the correct medical entities.
\end{inparaenum}

The scale runs from 1 to 5. The higher the number, the better. For the fluency metric, the ratings refer to incomprehensible, disfluent, non-native, good and flawless English, respectively. Similarly, for the adequacy metric these correspond to none, little meaning, much meaning, most meaning and all meaning, respectively. The ratings from the various annotators are averaged and shown in Table \ref{tab:h_results}. We compute the Fleiss’ kappa \cite{fleiss1971measuring} score to measure the inter-annotator agreement.

\section{Results and Analysis}
\label{sec:result}
Table \ref{tab:auto_results} and Table \ref{tab:h_results} show the automatic and human evaluation results of baselines and the proposed models.

\subsection{Automatic Evaluation}
Table \ref{tab:auto_results} shows the results using automatic evaluation metrics on the \textit{CDialog} dataset. {On most metrics, we see that BioBERT-Entity outperforms Bert-Entity and BART-entity models\footnote{{We did a t-test \cite{lehmann2006testing} with the null hypothesis between proposed (BioBERT-Entity) and best baseline(BioBERT) (and BART and BERT with and without entity). For both settings the p-value was less than 0.001, indicating that the proposed methods significantly outperform the baselines.}}}, demonstrating the effectiveness of incorporating medical entities with biomedical embeddings as additional learning signals for improving the task of medical dialog generation. Overall, we observe that entity based models tends to perform better and capture majority of the entities present in the dialog. On CDialog, BioBERT-Entity yields a significant performance improvement by a margin of around 12.25\% in F1 score, and 52.94\% in BLEU-4 on the test set when compared to the strongest baseline, BioBERT. Apart from word overlapping based metrics, we also notice significant improvement in embedding based metrics denoting efficient decoding using relevant entity information. Comparison to more baseline models can be found in Appendix \ref{sec:auto_r}.

\subsection{Human Evaluation Results}
Table \ref{tab:h_results} shows the result of human evaluation. Entity based models outperform the baseline models on fluency, adequacy, and medical entity relevance, demonstrating consistency with automatic evaluation results. All of the kappa values are greater than 0.75, indicating that the annotators are in good agreement.

In Table \ref{tab:gen_resp}, we present a few example conversations as predicted by the entity based \textit{BioBERT-Entity}, \textit{BART-Entity} and vanilla  \textit{BioBERT} and \textit{BART} models on the test set from CDialog corpus. As seen in the first example, BioBERT-Entity correctly decodes the response by utilising the context information and provides counselling to the patient. In the same example, we may note that BioBERT-Entity, as opposed to models without entity information like BioBERT, generates a more adequate response by utilizing the entity \textit{"diabetes"}.

\begin{table*}[ht!]
\small
\begin{center}

\begin{adjustbox}{max width=1.0\textwidth}
\renewcommand{\arraystretch}{1.2}
\setlength\tabcolsep{1.2pt}
\begin{tabular}{|>{\centering\arraybackslash}p{2.3cm}|p{18.5cm}|}
\hline
 & \textbf{Conversation}
\\
 \hline

\textbf{Patient} & hello doctor, i have {type 1 diabetes}, does that mean i can't get the virus just mildly ? am i likely to have severe complications ? \textbf{Symptom}: {diabetes} \\
\textbf{Doctor} & {diabetes + covid-19}. because you are a {type 1 diabetic} it means that you are at greater risk for complications if you contract {covid-19}. \textbf{Symptom}:\textit{diabetes}, \textbf{Disease}:\textit{Covid-19}
\\
\hline
\textbf{BioBERT-Entity} & yes, it doesn't mean diabetic. but most cases of diabetes are painless. they grow very rapidly. don't change, however they have some risk and the complications are serious.	\\

\textbf{BioBERT} & any other medical causes should be considered quickly. i understand your situation and would try to help. \\

\textbf{BART-Entity} & most likely you are having a diabetes disease. \\

\textbf{BART} &	have you consulted any physician ? \\
\hline
\textbf{Patient} & hi doctor, my 5 year old son woke up not feeling well. i measured his temperature and mouth reading is 38.3 and armpit is 38.2.  \textbf{Symptoms}:\textit{fever},  \textbf{Disease}:\textit{Covid-19},
\textbf{Other aspect}:\textit{age}
\\
\textbf{Doctor} & what more symptoms does he show ?\\

\textbf{Patient} & he says his head is sore and his neck is sore. am scared it may be covid - 19 ? \textbf{Symptoms}:\textit{head sore , neck sore}, \textbf{Disease}: \textit{covid - 19}
\\
\textbf{Doctor} & arrange testing. stay home, provide fluids and ibuprofen after checking his temperature.  \textbf{Symptoms}:\textit{fever}, \textbf{Medications: }\textit{ibuprofen}, \textbf{Remedies: }\textit{stay home , provide fluids} \\
\textbf{Patient} &  what to do if he does not get well ? \\
\textbf{Doctor} & if he does not look well to you, please arrange a virtual or in person appointment with your pediatrician.\\
\textbf{Patient} & are there any precautions i should follow ?
\\
\hline
\textbf{Doctor} & avoid contact with others, use good hand washing and consider testing for influenza and coronavirus. would you like to video or text chat with me? \textbf{Medical Test: } \textit{influenza test, covid test}, \textbf{Remedies: }\textit{use hand sanitizer}
\\
\hline
\textbf{BioBERT-Entity} & ok, you can send a picture without the difference. \\

\textbf{BioBERT} & no problem, glad to be of help . be safe and avoid hand washing or dusty hands. \\
\textbf{BART-Entity} & 	 please call your doctor as soon as possible. if he develops some signs of covid - 19, he should be examined and tested as soon as possible.	  \\

\textbf{BART} & your welcome \\
\hline
\end{tabular}
\end{adjustbox}
    
\end{center}
\caption{
\label{tab:gen_resp} Case Study: Examples of predictions from our proposed models on the test set. We attempt to predict Doctor's responses based on the sequence of Patient-Doctor-Patient utterances. The corresponding sets of medical entities are bolded.}

\end{table*}

\subsection{Error Analysis} 
\label{sec:error_analysis}
Using the generated responses, we conduct a thorough examination of our proposed model and classify the errors it encounters into the following categories:

\begin{inparaenum}
\item \textbf{Generic Response}: We see cases of generic responses by the doctor, such as \textit{``would you like to video or text chat with me?"} in about 4\% of our predicted responses by all the models leading to reduced medical entity presence. The reason being that there for many samples in the training data where such responses are present to maintain proper information flow and leading to a reasonable conversation. 

\item \textbf{Non-Fluency}: We observe around 5\% cases of non fluency such as ``if you were feeling ?", ``yes, we can think you give me?" mostly for BERT and BART models. The reason for this is that these models do not take into account medical entities because they are not trained on biomedical data, which leads to inconsistency in responses since they miss important medical terms while predicting responses.

\item \textbf{Inadequacy}: The model sometimes fails to predict correct responses for patient utterances having a large set of context utterances. For example in Table \ref{tab:gen_resp}, we may observe in the second sample that since the conversation history comprises of more than six utterances. The model fails to keep track of the previous information and hence generates an inadequate or a generic response.

\item \textbf{Incorrect entity prediction}: In around 10\% cases, the model predicts some irrelevant medical entities resulting in contextually incorrect responses. For example, \textbf{Patient}:\textit{ i am experiencing nasal congestion, sneezing (unaffected by: recent exposure to allergens, exposure to secondhand smoke), sore throat, itchy eyes, ear pressure, nasal drainage, post nasal drip, eye irritation, runny nose, and watery eyes}; \textbf{Doctor}: \textit{i think it is itching/congestion. with the itching could be seasonal allergies would consider benadryl 1/2 to 1 tab at bedtime and zyrtec during the day. itching is pretty specific for allergies?}; \textbf{Predicted Response:} \textit{hi, also called urti-allergy. have you taken any medicines?} As can be seen, the predicted response missed all of the entities mentioned in the patient's utterance. However, the reason could be that because many entities were mentioned in the utterance, the model was confused and mentioned "urti-allergy" which is also very close to the mentioned symptoms.

\end{inparaenum} 

More details on the performance of baseline models on these errors can be found in Appendix \ref{sec:error_analysis_extended}.

\section{Conclusion}
In this paper, we have created an enriched multi-turn medical dialog corpora with manually labeled medical entities. The dataset is typically constructed for the purpose of developing an efficient medical dialog system, with an average dialog length of 8. To facilitate effective conversation understanding and generation, we propose an entity-aware neural conversational model for medical dialog generation. The evaluation results on two benchmark medical datasets show that a BERT-based model with biomedical embeddings and relevant medical entities can successfully generate correct and informative responses. 

In the future, we aim to use a medical knowledge graph generated using a UMLS database to provide domain knowledge into medical dialogues and model the relationship between different medical entities. The codes and dataset used to replicate our findings are available at \hyperlink{https://github.com/deekshaVarshney/CDialog}{https://github.com/deekshaVarshney/CDialog}; 

\section{Ethical Declaration}
All of the datasets used in this study are freely available to the public which are collected from public websites. We followed the policies for using those data and did not violate any copyright issues. The dataset used in this paper is solely for academic research purposes. In a real-world application, medical dialogue systems could be used to counsel patients and collect data for diagnosis. Even if the agent makes a few minor mistakes during the process, doctors will eventually take over in the end. Annotation was done by a dedicated team of people who work full-time. { Dataset is medically verified by the health department of our institute. We are not disturbing any health related information and only adding generic statements in order to maintain the flow of the conversation. We further got the data collection and annotation process reviewed by our university review board.}



\section{Limitations}
Detailed cases of limitations by our model is described in Section \ref{sec:error_analysis}. Modelling medical entities is a challenging task in dialog generation. We aim to further investigate this task in the future.

\section{Acknowledgement}
We would like to thank the reviewers for their constructive comments. 
Authors gratefully acknowledge the support from the projects ``Percuro-A Holistic Solution for Text Mining``, sponsored by Wipro Ltd; and ``Sevak-An Intelligent Indian Language Chabot``, sponsored by Imprint 2, SERB, Government of India.


\bibliography{anthology,custom}
\bibliographystyle{acl_natbib}

\appendix

\section{Dataset statistics}
\label{sec:data_details}
Table \ref{tab:dataset_stats} presents the dataset statistics for the proposed CDialog dataset. The dataset is split into 80:10:10 ratio for preparing the training, test and validation sets.

We conduct several experiment to show the effectiveness of the annotation of entities. They are described as follows. Since, we have broken the longer utterances into short utterances, having extra information in the form of entity annotation is clearly useful. This is already demonstrated by our experiments in Table \ref{tab:auto_results}, by building models both with and without entities. The results clearly show improvement in performance for models with entity. Similarly, we conduct an additional experiment with the Ext-CovidDialog dataset and observed that with the entities there is no improvement in the model. Hence, showing that for shorter utterances the entity annotation is more useful.
Results on Ext-CovidDialog:
\textbf{BioBERT} - \textit{F1-score}:  0.222;
\textbf{BioBERT + Entity} - \textit{F1-score}:  0.211

\begin{table}[ht!]
\begin{center}
\renewcommand{\arraystretch}{1.2}
\setlength\tabcolsep{1.3pt}
\begin{adjustbox}{max width=0.50\textwidth}
\centering
\small
\begin{tabular}{|c|c|c|c|c|c|c|}
\hline
\textbf{Statistics}            & \multicolumn{3}{c|}{\textbf{CDialog}}    \\ \hline
\textbf{\#Conversations}          & \multicolumn{3}{c|}{1,012}  \\ \hline
\textbf{\#Utterances}          & \multicolumn{3}{c|}{7,982}  \\ \hline
\textbf{\#Tokens}              & \multicolumn{3}{c|}{1,085,204} \\ \hline
\textbf{Average \# Utterances} & \multicolumn{3}{c|}{8}   \\ \hline
\textbf{Maximum \# Utterances} & \multicolumn{3}{c|}{48}     \\ \hline
\textbf{Minimum \# Utterances} & \multicolumn{3}{c|}{2}   \\ \hline
\textbf{Average \# Tokens}     & \multicolumn{3}{c|}{136}  \\ \hline
\textbf{Maximum \# Tokens}     & \multicolumn{3}{c|}{5,313}    \\ \hline
\textbf{Minimum \# Tokens}     & \multicolumn{3}{c|}{2}    \\ \hline
\end{tabular}
\end{adjustbox}
\end{center}
\caption{Dataset statistics}
\label{tab:dataset_stats}
\end{table}

\section{Annotation Details}

\label{sec:annotation_guide}
\begin{figure}[ht!]
    \centering
    \includegraphics[width=0.40\textwidth]{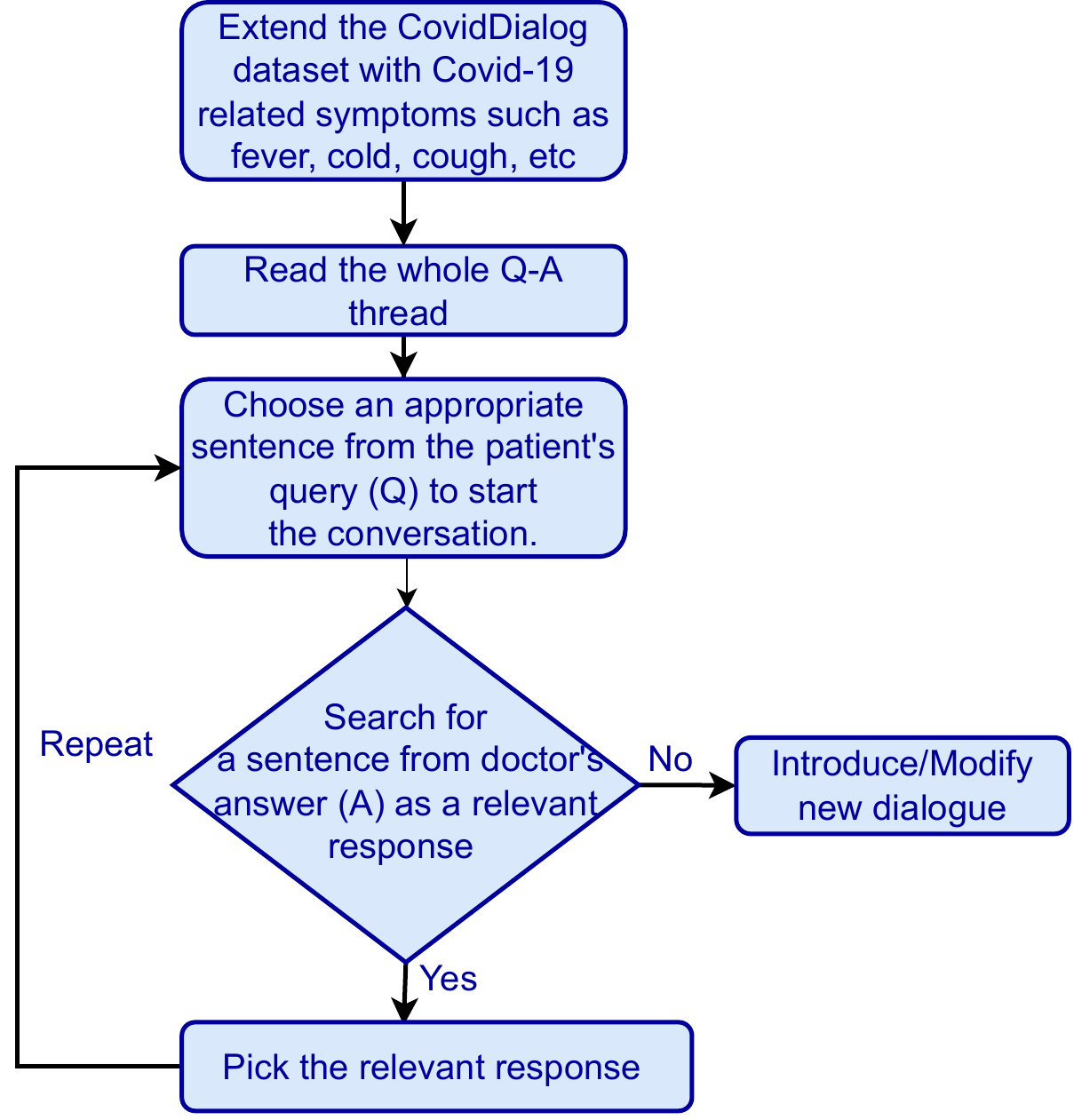}
    \caption{Construction and annotation pipeline of CDialog Dataset}
    \label{fig:flowchart}
\end{figure}

\paragraph{Annotation Guideline:}
{Given a query from patient and an answer from doctor, the task is to convert it into a multi-turn dialog by selecting sentences from the query-answer pair such that  they form a sensible multi-turn conversation. Each turn in the conversation contains an utterance by the patient and a response by the doctor. Figure \ref{fig:flowchart}, shows an overview of the pipeline for creating the multi-turn dialog data.}

\begin{enumerate}
    \item {For each sample query-answer pair, we employ two annotators, one who produces utterances for the patient and one who acts as a doctor and selects relevant sentences as responses. This configuration has several advantages over using a single annotator to serve as both a patient and a doctor such as when two annotators chat about a passage, their dialogue flow is natural and when one annotator responds with a vague response, the other can raise a flag, which we use to identify bad workers.
    \item Both the acting patient and doctor sees the original query and answer and also the conversation that happened until now i.e utterances and response from previous turns.
    \item While framing a new utterance for starting the conversation, we want annotators to see the longer query and mostly pick the first sentence as their utterance and modify accordingly to begin the conversation. For example, as shown in Figure \ref{fig:example}, the annotator picks the " I am a 23-year-old man" sentence from $Q$ and adds "and I have some queries regarding coronavirus. Can you help me?" in order to start the conversation.
    \item While responding, we want the annotator to look into the longer answer (c.f. $A$ in Figure \ref{fig:example}) and pick the appropriate sentence as the doctor's utterance and we further ask them to sometime respond with only generic sentences such as \textit{Is there anything else you wanna tell?} (c.f $X_{12}$), \textit{Yes sure, please state your concern.} (c.f $X_2$) to generate a natural conversation. 
    \item For medical entity annotation, seven empty columns are provided to choose the relevant medical term for the different categories as defined in Section \ref{sec:const_det}. For example in Figure \ref{fig:example}, for utterances $X_4$, the relevant medical entities to be annotated are \textit{Symptom}: Anxiety; \textit{Disease}: Covid-19. The annotators were also asked to remove any names to anonymize the data.}
\end{enumerate}

\paragraph{Annotators details:} {The annotators are regular employees (paid monthly as per university norms) at the rate of 35k/month. The annotators have been employed in our research group and they have been working on similar projects since the last three years.}

\section{Implementation Details}
\label{sec:imp_det}
All the experiments are implemented using Pytorch framework.  BART and BioBERT had hidden size of 1024 while BERT had hidden size of 512. The number of layers is set to 2, 12 and 6 for BERT, BART and BioBERT model respectively. For all the three model BERT, BART and BioBERT number of parameters were 96764928, 457762816 and 360749056 respectively. We use grid search to get the optimal hyperparameter values. We use the AdamW optimizer with learning rate fixed to 0.0005 and the beam size set to 1, while decoding the responses. We choose the best model when the loss on the validation set does not decrease further. We use the GeForce GTX 1080 Ti as the computing infrastructure. Each model is trained up to 30 epochs. After three runs with different random seeds for each method, the variances of the results are at most 1e-4, and they have no impact on the trend.

\section{Results}
\subsection{Automatic Evaluation}
\label{sec:auto_r}
{We also compare our proposed approaches with LSTM based state-of-the-art models such as Seq2Seq \cite{vinyals2015neural}, HRED \cite{serban2015hierarchical} and VHRED \cite{serban2017hierarchical}. Seq2Seq obtains a F1-score of 5.20 and BLEU-4 score of 0.001 on test set of our proposed CDialog dataset. HRED obtains a F1-score of 5.67 and a BLEU-4 score of 0.003 with an embedding average, extrema and greedy score of 0.611, 0.302, 0.542 respectively. VHRED obtains F1-score of 6.11 and a BLEU-4 score of 0.003 with an embedding average, extrema and greedy score of 0.621, 0.304, 0.552 respectively.}

\section{Error Analysis} 
\label{sec:error_analysis_extended}
Performance of baseline models on Inadequacy and Incorrect entity prediction.
\begin{inparaenum}

\item \textbf{Inadequacy}:  {The prediction by baseline models BART and BioBERT models is shown in Table \ref{tab:gen_resp}. As can be seen, the baseline models also struggle to maintain track of information, resulting in an insufficient or generic response.}

\item \textbf{Incorrect entity prediction}: {For the example shown in 
\ref{sec:error_analysis}, 4-th point, the performance of baseline models is as follows:
\textit{BERT}: have you been recently? please send for any more information. i have read your query in detail.
\textit{BART}: do you have family history?
\textit{BioBERT}: not allergy. if you have already taken antibiotics, it may help. did you have any other contact with a doctor? It can be noted that the baseline models perform even worse than the models with entities in terms of retaining relevant clinical information in the predicted response.}
\end{inparaenum}

\end{document}